\documentclass[10pt,twocolumn,letterpaper]{article}

\usepackage[pagenumbers]{cvpr} 

\usepackage[dvipsnames]{xcolor}

\definecolor{cvprblue}{rgb}{0.21,0.49,0.74}
\usepackage[pagebackref,breaklinks,colorlinks,citecolor=cvprblue]{hyperref}

\usepackage{multirow}
\newcommand{\tabincell}[2]{\begin{tabular}{@{}#1@{}}#2\end{tabular}}
\usepackage{pifont}
\newcommand{\yes}{\text{\ding{51}}}
\newcommand{\no}{\text{\ding{55}}}
\usepackage{bm}

\usepackage{algorithm,algpseudocode}

\title{Tracking with Human-Intent Reasoning}

\author{Jiawen Zhu$^{1,3}$, \ Zhi-Qi Cheng$^{2,\dagger}$, Jun-Yan He$^3$, Chenyang Li$^3$, \\
	Bin Luo$^3$, Huchuan Lu$^{1,\dagger}$, 
	Yifeng Geng$^{3}$, Xuansong Xie$^{3}$ \\
	$^1$Dalian University of Technology \quad $^2$Carnegie Mellon University\\
	$^3$Institute for Intelligent Computing, Alibaba Group 
}

\begin{document}
\maketitle
\begin{abstract}
Advances in perception modeling have significantly improved the performance of object tracking.
However, the current methods for specifying the target object 
in the initial frame are either by {1}) using a box or mask template, or by {2}) providing an explicit language description.
These manners are cumbersome
and 
do not allow the tracker to have self-reasoning ability.
Therefore, this work proposes a new tracking task --- \textbf{Instruction Tracking}, which involves providing implicit tracking instructions that require the trackers to perform 
tracking automatically in video frames.
To achieve this, we investigate the integration of knowledge and reasoning capabilities from a Large Vision-Language Model (LVLM) for object tracking.
Specifically, we propose a tracker called \textbf{TrackGPT}, which is capable of performing complex reasoning-based tracking. 
TrackGPT first uses LVLM to understand tracking instructions and condense the cues of what target to track into referring embeddings.
The perception component then generates the tracking results based on the embeddings.
To evaluate the performance of TrackGPT, we construct an instruction tracking benchmark called \textbf{InsTrack}, which contains over one thousand instruction-video pairs for instruction 
tuning and evaluation.
Experiments show that TrackGPT achieves competitive performance on referring video object segmentation benchmarks, such as getting a new state-of-the-art performance of 66.5 $\mathcal{J}\&\mathcal{F}$ on Refer-DAVIS.
It also demonstrates a superior performance of instruction tracking under new evaluation protocols. 
The code and models are available at \href{https://github.com/jiawen-zhu/TrackGPT}{https://github.com/jiawen-zhu/TrackGPT}.
\end{abstract}    
\newcommand\blfootnote[1]{%
	\begingroup 
	\renewcommand\thefootnote{}\footnote{#1}%
	\addtocounter{footnote}{-1}%
	\endgroup 
}
{
	\noindent \blfootnote{\hspace{-6mm} 
		$^\dagger$ Corresponding authors.
	}
}
\section{Introduction}
\label{sec:intro}

\begin{figure}[h]
	\centering
	\includegraphics[width=0.485\textwidth]{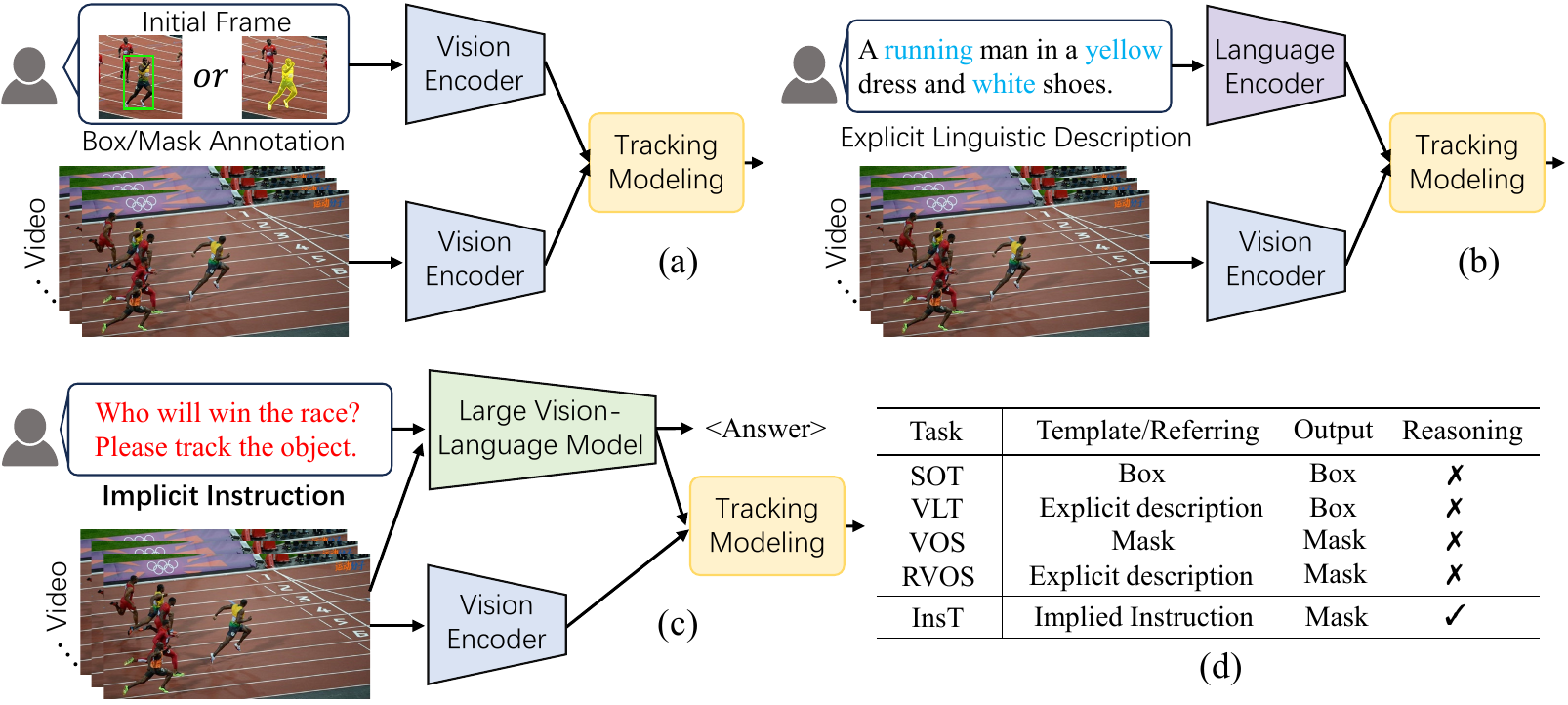}
	\vspace{-5mm}
	\caption{ Comparison of different tracking paradigms. (a) Tracking by box/mask template. (b) Tracking by explicit linguistic description. (c) Tracking by human instruction.
		(d) Differences in existing tracking tasks. SOT: single object tracking. VLT: vision-language tracking. VOS: video object segmentation. RVOS: referring video object segmentation. InsT: instruction tracking.}
	\vspace{-3mm}
	\label{fig:abs}
\end{figure}

Object tracking is a task where an algorithm continuously estimates the shape and position of a specific object of interest in a video.
To begin tracking, the algorithm requires a clear indication of the target object. There are two primary ways of specifying the target object in existing tracking paradigms:~1) \emph{Direct bounding box template}~\cite{siameserpn,dimp,transt,ostrack,mixformer} or \emph{segmentation mask annotations}~\cite{stm,cfbi,aot,xmem}.~2) \emph{Explicit linguistic description}~\cite{tnl2k,jointnlt,referformer,onlinerefer,r2vos}. The bounding box and segmentation mask directly provide information about the target object's visual attributes
in the initial frame (Fig.~\ref{fig:abs}(a)). On the other hand, the {explicit linguistic description} provides a detailed description of the object's appearance and behavior features, which must refer to a specific target and be ensured by the annotator (Fig.~\ref{fig:abs}(b)).

Despite progress in perceptual modeling~\cite{alexnet,resnet,vit,bert} has allowed trackers to accurately extract the corresponding visual and linguistic features for high-performance object tracking, existing trackers still heavily rely on 
direct visual templates or explicit language descriptions to specify targets in real-world scenarios involving human-robot systems and interactive embodied agents.
However, these manners for specifying the target object can be {impractical and cumbersome}, especially when annotating fine-grained masks.
Furthermore, it can also be challenging to \emph{represent the target object in a way that ensures the tracker can locate it accurately}, even within a limited amount of time.

Ideally, a more practical approach to human-tracker interaction would involve {implicit instruction} to identify target objects, as demonstrated in Fig.~\ref{fig:abs}(c). The goal is to develop a tracker that can \emph{autonomously interpret human cues} and adapt to \emph{implicit instruction} and {preferences learned through observation over time}. 
Such \emph{self-reasoning capability} is crucial for the advancement of the next generation of intelligent sensing. However, a major challenge lies in the fact that current systems still lack proficiency in discerning {human instructions}. The \emph{crux of the matter} is that \emph{subtle visual and contextual cues} that are intuitively understood by humans remain difficult for trackers to grasp.

To overcome inherent limitations and advance object tracking with more intelligence, we propose a new tracking task called \emph{instruction tracking}.
Compared to vanilla object tracking, the instruction in this task is a more implicit query, such as "\emph{Who will win the race?}".
This type of instruction aligns better with the way humans ask questions. 
It also implies that the tracking system needs to combine language instructions with visual tracking scenes to reason, understand the elements in the video, and locate the corresponding object of interest for subsequent tracking.

To address the challenge of instruction tracking, we introduce \emph{TrackGPT}, a tracker with the ability to reason human intent. 
To enable the tracker to understand human instructions, we utilize a large vision-language model (LVLM) to process the instruction and visual input of the initial frame, and generate referring cues that apply to all subsequent frames for target object tracking.
Typically, the decoder in the perception component decodes and tracks the object and generates mask outputs. 
In addition, we propose a rethinking mechanism based on purport scores for LVLM to evaluate whether the current results meet the instruction purport.
To further improve the robustness and accuracy of instruction tracking, we design a cross-frame referring propagation module to exploit the position prior to the object and adjust the appearance change in a video.

Moreover, we develop a benchmark called \emph{InsTrack} to assess the effectiveness of instruction tracking.
It requires the tracker to have the ability to comprehend human instructions and track the target object correctly. 
The benchmark comprises over a thousand instruction-video pairs for tuning and evaluation purposes.
Our TrackGPT is capable of understanding human instructions and achieving robust tracking.
It also delivers competitive performance on conventional referring video object segmentation benchmarks like Refer-Youtube-VOS~\cite{refer-youtube-vos} and sets a new state-of-the-art on Refer-DAVIS$_{17}$~\cite{refer-davis}. After instruction tuning on the InsTrack train set, our TrackGPT-7B and TrackGPT-13B models further enhance instruction tracking performance. 
Our contributions are summarized as:
\begin{itemize}[leftmargin=*]
	\item{We introduce a new task called instruction tracking, where a tracker must have the self-reasoning capability, autonomously interpret implicit instruction, and track the target object.
		We also constructed a benchmark, InsTrack, 
		for instruction tracking tuning and evaluation.}
	\item{We propose TrackGPT, the first tracker that can comprehend human intent by leveraging the reasoning capability of LVLM. Besides, a rethinking mechanism is proposed for self-correcting the prediction that deviates from the instruction purport, and a cross-frame referring propagation module is designed for utilizing adjacent-frame cues to achieve robust and accurate tracking.}
	\item{The experimental results show that TrackGPT achieves comparable performance in referring video object segmentation and exceptional instruction tracking ability.}
\end{itemize}

\section{Related Work}
\label{sec:formatting}

\begin{figure*}[t]
    \centering
    \vspace{-2mm}
	\includegraphics[width=0.92\textwidth]{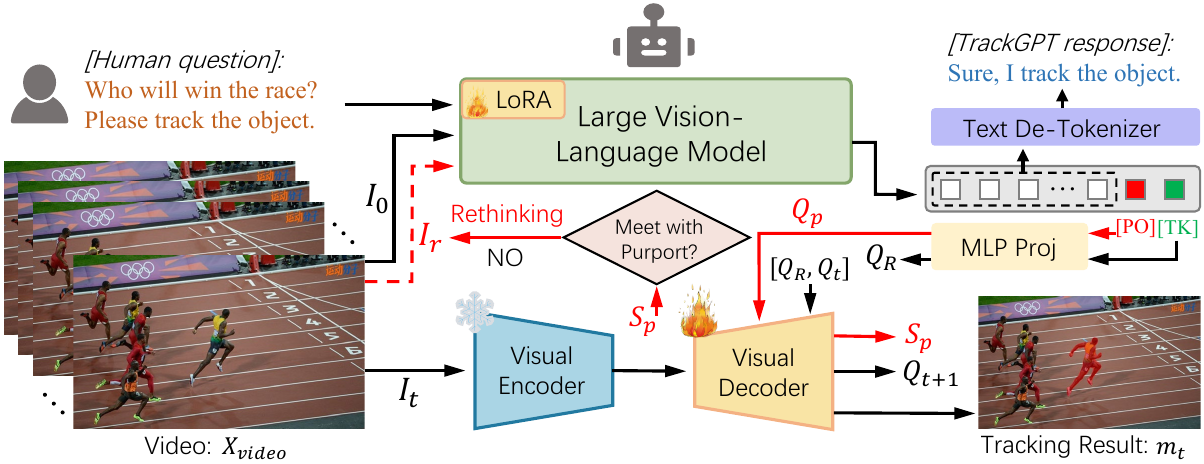}
	\vspace{-3mm}
	\caption{Overview architecture of TrackGPT. The current frame $\bm{I}_{t}$ is fed into a visual encoder to extract visual features. The initial frame $\bm{I}_0$ and the corresponding tracking instruction are sent into an LVLM brain to comprehend human intent and generate the referring queries $\bm{Q}_R$ for the target object. Finally, the decoder receives the visual features and linguistic embeddings, predicting the tracking results $\bm{m}_t$.
	The red arrows in pipeline indicate the proposed rethinking mechanism, and $\bm{Q}_t\rightarrow\bm{Q}_{t+1}$ represents the cross-frame referring propagation.
}
	\label{fig:overview}
\end{figure*}

\subsection{Video Object Tracking}
Single object tracking (SOT) methods~\cite{siamesefc,siameserpn,dimp,transt,ostrack} are given the initial bounding box of the target object, and the trackers directly extract the visual features of the template using deep neural networks~\cite{alexnet, resnet, vit}, and then predict the bounding box tracking results.
Due to the directly accessible visual features of the target object, mainstream tracking architectures commonly take tracking modeling as a matching problem.
For instance, TransT~\cite{transt} proposes the transformer-based fusion module to composite the template and search region features for object localization.
Providing a segmentation mask template, the tracking modeling falls into the problem of semi-supervised video object segmentation (VOS).
VOS methods~\cite{stm,cfbi,aot,xmem} obtain the finer-grained mask template than the bounding box in the initial frame, and the tracking results in subsequent frames are obtained through the propagation of the initial mask.
The typical approach STM~\cite{stm} treats past frame predictions as the memory for current frame matching, and subsequent methods~\cite{cfbi,aot,xmem} have iteratively improved upon it.

When visual templates are unavailable instead a textual description is given, the problem transforms into referring object tracking.
According to different output formats, these methods can be divided into vision-language tracking (VLT)~\cite{tnl2k,vlt,jointnlt} and referring video object segmentation (RVOS)~\cite{yofo,referformer, r2vos, onlinerefer}, which respectively output bounding box and segmentation mask results.
This method typically employs a BERT-based~\cite{bert,roberta} text encoder to extract feature embeddings from textual descriptions, and then perform grounding modeling with the features extracted by the visual encoder.
To achieve accurate referring segmentation, some methods~\cite{hui2021collaborative,lbdt} design spatial-temporal modeling modules to explore additional temporal cues.

The above methods effectively extract visual templates or linguistic referring information and adopt well-designed object modeling modules to track the target object in videos.
However, the template or referring information they rely on demands excessive human effort. 
In real-world applications, it is hard to provide a bounding box, segmentation mask, or detailed language description for tracking.

\subsection{Large vision-language Model}
Large Language Models (LLMs)~\cite{llama,gpt3.5} have demonstrated astonishing reasoning capabilities to comprehend natural language, and efforts have been made to explore how vision tasks can benefit from them.
To inject visual information into LLMs, BLIP-2~\cite{blip2} employs a Q-Former to project the features from the vision transformer to a text decoder.
MiniGPT-4~\cite{minigpt4} connects the vision features with a Vicuna~\cite{vicuna} model by a linear layer. 
Similarly, mPLUG-OWL~\cite{mplug-owl} uses a visual abstractor layer to summarize visual information into several tokens, which are then combined with text embeddings and fed into a pre-trained LLM.
LLaVA~\cite{llava} performs a two-stage fine-tuning by image-text feature alignment and instruction tuning.
While these methods have explored vision-language tasks, their focus is primarily on text generation rather than vision outputs.
More recently, some works (e.g., \cite{visionllm, kosmos2, gpt4roi,detgpt}) explore richer forms of vision-language interaction, enabling the completion of a wider range of vision-centric tasks.
For example, DetGPT~\cite{detgpt} accomplishes instruction-based detection by integrating an open-vocabulary detector with a multi-modal LLM.
LISA~\cite{lisa} achieves reasoning-guided segmentation by employing a multi-modal LLM and mask decoder.
These works are primarily focusing on some fundamental image tasks (\eg, VQA, detection, and segmentation).
In this work, we are dedicated to integrating the reasoning abilities of LVLM into the field of video object tracking.
It has application value for extending to downstream tasks and is also a further exploration of the future of more intelligent and interactive perceptual technologies.

\section{Methodology}

\subsection{Instruction Tracking Task}
Given an input video $\bm{X}_{video}$ and corresponding object prompt $\bm{X}_{prompt}$, the object tracking task is aiming to learn a tracker $\mathcal{F}_{tracker}:  \{\bm{X}_{video}, \bm{X}_{prompt}\} \rightarrow \bm{Y}$ to predict the shape and position of the target object.
$\bm{X}_{prompt} \in \{\bm{X}_{box}, \bm{X}_{mask}, \bm{X}_{text}\}$ means the object prompt formulation can be a bounding box template, segmentation mask template, or language description.
$\bm{Y} \in \{\mathcal{B}, \mathcal{M}\}$ denotes the output which can be a bounding box or mask.

Conventional object tracking heavily relies on direct visual templates or explicit language descriptions to specify targets in given videos, which can be impractical and cumbersome.
Unlike traditional object tracking, instruction tracking faces more significant challenges as its implicit tracking instructions, represented by $\bm{X}_{inst}$.
The implicit instructions align better with the way humans ask questions. 
In instruction tracking, the tracker must comprehend human instruction, determine the target object in the video, and complete the tracking process without relying on explicit cues that directly describe objects.
Formally, the objective of instruction tracking is to predict object masks from input video frames: $\{\bm{X}_{video}, \bm{X}_{inst}\} \rightarrow \mathcal{M}$.

\subsection{Overview of TrackGPT}
The overall architecture of TrackGPT is presented in Fig.~\ref{fig:overview}. TrackGPT mainly consists of an LVLM brain, a pre-trained visual encoder, and a query-based mask decoder.
To effectively understand the tracking instruction from humans, we incorporate an LVLM to receive the initial frame $\bm{I}_0$ and the corresponding instruction $\bm{X}_{inst}$.
The referring cues of the target object will be generated in the form of language token $\langle \text{TK} \rangle$, TrackGPT will send it to the perception component while completing the dialog with human.
On the other hand, the vision encoder specializes in the extraction of the visual features and feeds them into a mask decoder which accurately segments the target object, prompted by the referring embeddings provided by the LVLM reasoning brain.

\subsection{LVLM Reasoning Brain}
\label{sec:trackgpt}

The LVLM reasoning brain $\mathcal{F}_{Brain}$ is constructed by the LVLM model, MLP projection layers, and a text de-tokenizer, respectively.
As the core reasoning component of the tracking system, it is responsible for understanding human intent based on multi-modal vision and instruction inputs, automatically locating the target object of interest, and coordinating the perception component to accomplish online object tracking.
Formally, the input and output are formed as follows:
\begin{align}
	\label{eq:lvlm_brain}
	\mathcal{T}, \bm{Q}_{R}, \bm{Q}_{p}= \mathcal{F}_{Brain}(\bm{I}_{R}, \bm{X}_{inst}),
\end{align}
where $\mathcal{T}$ is the text feedback from the LVLM model,
$\bm{Q}_{R}$ and $\bm{Q}_{p}$ are the referring and purport queries, and $\bm{X}_{inst}$ is the instruction that provided by the human.
$\bm{I}_{R}$ denotes the reference frame, which can be the initial frame or the following frame determined by the rethinking mechanism. 

\noindent\textbf{Large Vision-Language Model.}
To understand the implicit instruction from humans, we incorporate the idea of LVLMs~\cite{minigpt4, mplug-owl, llava}.  
We adopt an LVLM~(\ie, LLaVA~\cite{llava} in our case) to simultaneously receive tracking instructions and produce the tokens for the tracking. Technically, the tokens are generated by 
$\mathcal{S}=\mathcal{F}_{LVLM}(\bm{I}_{R},\bm{X}_{inst})$,
where $\mathcal{S}=\{\langle\text{T}_0\rangle, \langle \text{T}_1 \rangle, \dots, \langle \text{T}_N \rangle, \langle \text{PO} \rangle,\langle \text{TK} \rangle\}$
is the token vector predicted by the original LVLM model $\mathcal{F}_{LVLM}$. 
Following LISA~\cite{lisa}, TrackGPT obtains the understanding of the target object by expanding the LVLM vocabulary with two extra new tokens $\langle \text{PO} \rangle$ and $\langle \text{TK} \rangle$, which represent the purport and referring token respectively.
To fine-tune the LVLM model effectively and efficiently, we incorporated the LoRA~\cite{lora} technique by adding a small scale, low-rank decomposed, weight ${W}_{lora}$ to the basic LVLM, tuning this weight to empower the LVLM with the tracking capability, consuming little extra computational cost.

\noindent \textbf{MLP Projection \& De-Tokenizer.} 
After the  $\mathcal{S}$ is produced, the $\langle \text{TK} \rangle$ token is projected into referring embeddings $\bm{Q}_R$ by a 3-layer MLP projection function $\mathcal{\phi}(\cdot)$, which transforms the dimension from $\langle \text{TK} \rangle \in R^{4096}$ to $\bm{Q}_R \in R^{256}$ to match the input dimension of visual decoder. Therefore, $\bm{Q}_R$ can be sent to the decoder and serves as an important contextual cue for subsequent tracking modeling. 
Similarly, the purport query $\bm{Q}_{p}=\mathcal{\phi(\langle \text{PO} \rangle )}$ is also generated in the same way. Moreover, the rest text tokens are fed into a text de-tokenizer to synthesize the text feedback $\mathcal{T}$.
\begin{algorithm}[!t]
	\footnotesize
	\caption{\footnotesize TrackGPT with Rethinking Mechanism}
	\label{alg:rethinking}
	\begin{algorithmic}[1]
		\State \textbf{Input:} video frames $\bm{X}_{video}\!\!\!=\!\!\!\{\bm{I}_0,...,\bm{I}_n\}$, tracking instruction $\bm{X}_{inst}$, LVLM brain $\mathcal{F}_{Brain}$, and perception component $\mathcal{F}_{Percep}$;
		\State \textbf{Output:} tracker answer $\mathcal{T}$, video mask $\mathcal{M}=[\bm{m}_0,...,\bm{m}_n]$;
		\State \textbf{For} each frame $\bm{I}_{t}$ in $\bm{X}_{video}$ \textbf{do}
		\State \hspace{\algorithmicindent} \textbf{if} $t == 0$ \textbf{then}
		
		\State \hspace{\algorithmicindent} \hspace{\algorithmicindent} LVLM reason: $\{\bm{I}_0, \bm{X}_{inst}\}\rightarrow\mathcal{T}, \bm{Q}_R, \bm{Q}_p$; \Comment{Eq. (\ref{eq:lvlm_brain})}
		\State \hspace{\algorithmicindent} \hspace{\algorithmicindent} Predict results:  $\{\bm{I}_0,\bm{Q}_R, \bm{Q}_p\}\rightarrow\bm{m}_{0},{S}_{p}$; \Comment{Eq. (\ref{eq:perception})}
		\State \hspace{\algorithmicindent} \textbf{else}
		\State \hspace{\algorithmicindent} \hspace{\algorithmicindent} \textbf{if} ${S}_{p} < \tau$  \textbf{then} 
		\State \hspace{\algorithmicindent} \hspace{\algorithmicindent}\hspace{\algorithmicindent} Rethinking: $\{\bm{I}_t, \bm{X}_{inst}\}\rightarrow\mathcal{T}, \bm{Q}_R, \bm{Q}_p$; \Comment{Eq. (\ref{eq:lvlm_brain})}
		\State \hspace{\algorithmicindent} \hspace{\algorithmicindent}\hspace{\algorithmicindent} Update queries: $\bm{Q}_R, \bm{Q}_p$;
		\State \hspace{\algorithmicindent} \hspace{\algorithmicindent} \textbf{end if}
		\State \hspace{\algorithmicindent} \hspace{\algorithmicindent}  
		Predict results:  $\{\bm{I}_t,\bm{Q}_R, \bm{Q}_p\}\rightarrow\bm{m}_{t},{S}_{p}$; \Comment{Eq. (\ref{eq:perception})}
		\State \hspace{\algorithmicindent} \textbf{end if}
		\State \hspace{\algorithmicindent}
		Update outputs:   $\mathcal{M}[i]=\bm{m}_{t}$;
		\State\textbf{end for}
		\State \textbf{return} $\mathcal{T},\mathcal{M}$;
	\end{algorithmic}
\end{algorithm}

\noindent\textbf{Rethinking Mechanism.}
In TrackGPT, if the LVLM brain only generates a referring query based on the initial frame image and the given instruction, it will lack the ability to react to changes in the video content.
The referring cues provided to the perception component will also be very limited.
To equip TrackGPT with the ability to respond to environmental changes,
we propose a rethinking mechanism as shown in Fig.~\ref{fig:overview}.
For ease of differentiation, we denote the relevant processes with red lines.
We introduce a concept of “purport” for instruction tracking, which refers to whether the current tracking results
align with the purport of the instruction.
To this end, we further expand the LVLM's vocabulary and introduce a purport token: $\langle \text{PO} \rangle$, 
which will be fed into the visual decoder to predict the purport score $\mathcal{S}_p$.
As the IoU score between the predicted result and the ground truth directly reflects the quality of the tracking results, we employ the IoU score as our purport score in TrackGPT.
The purport token continuously interacts with the visual perception component, when TrackGPT detects that the current tracking result no longer aligns with the purport ($\bm{S}_p<\tau$), it then performs rethinking in the next frame $\bm{I}_{r}$.
The image frame and language instruction are re-input into the LVLM to update the $\langle \text{TK} \rangle$ token.
Alg.~\ref{alg:rethinking} illustrates the working process of the rethinking mechanism during TrackGPT inference.
The rethinking mechanism enhances the performance of online tracking 
by reducing erroneous tracking that does not comply with the given instructions.
Moreover, some SOT methods~\cite{stark,mixformer} propose to train a score head to update the visual template for improving tracking reliability. 
Unlike them, the rethinking mechanism in TrackGPT introduces multi-modal vision-language communication and online updating, and does not require additional second-stage training for the score head.

\subsection{Perception Component}
\label{sec:perception}
The perception component $\mathcal{F}_{percep}$ consist of an image encoder $\mathcal{E(\cdot)}$ and a mask decoder $\mathcal{D(\cdot)}$:
\begin{align}
	\label{eq:perception}
	\bm{m}_{t}, S_{p} = \mathcal{F}_{percep}(\bm{I}_{t}, \bm{Q}_{R}, \bm{Q}_{p})
\end{align}
where $\bm{m}_{t}$ and $S_{p}$ denote the predicted mask of frame $\bm{I}_t$, and purport score, respectively. 

\noindent\textbf{Visual Encoder \& Decoder.}
The visual embedding $f_t$ is generated by 
$\bm{f}_t = \mathcal{E}(\bm{I}_{t})$, 
where $\mathcal{E}$ is an ViT encoder following the previous work~\cite{sam}. In the training phase, the encoder is frozen. To better estimate the quality of the mask prediction, we revamp the mask decoder~\cite{sam} in two aspects: 1) incorporating the prompt tokens with the referring token $\bm{Q}_{R}$; 2) fusing the purport token $\bm{Q}_{p}$ with the original iou token by multiplying. More details can be found in the {supplementary material}. Formally, the decoder process can be defined as:
\begin{align}
	\bm{m}_{t}, S_{p}, \bm{Q}_{t+1} = \mathcal{D}(\bm{f}_t, \bm{Q}_{R}, \bm{Q}_{t}, \bm{Q}_{p})
\end{align}
where $\bm{Q}_{t}$ and $\bm{Q}_{t+1}$ are the cross-frame online referring queries that will be adopted recurrently during tracking.
	
\begin{figure}[t]
	\centering
	\includegraphics[width=0.45\textwidth]{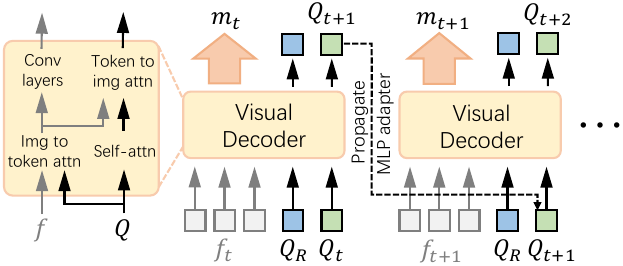}
	\vspace{-2mm}
	\caption{The proposed cross-frame referring propagation.
		An initial referring query $\bm{Q}_R$ and a cross-frame online referring query $\bm{Q}_t$ are responsible for decoding the target object mask.
	}
	\vspace{-2mm}
	\label{fig:prop}
\end{figure}
\noindent\textbf{Cross-frame Referring Propagation.}
The continuous changes in the object's appearance and the surrounding environment are a significant challenge in video object tracking. However, considering the heavy computational load of the LVLM, only the instruction-frame pair of the initial frame $\bm{I}_0$ or aforementioned rethinking frame $\bm{I}_r$ are sent into the LVLM to generate a $\langle \text{TK} \rangle$ token.
Almost all the frames utilize this $\langle \text{TK} \rangle$ token to complete the tracking modeling. However, in real-world tracking scenarios, the appearance and motion state of the target are usually changing constantly as time proceeds.  Therefore, using a fixed referring query from several frames limits the performance of the tracker in subsequent frames.
Taking into account the continuity of changes in the motion state of the target object, we propose to expand an online referring query $\bm{Q}_t$ that propagates across frames. As depicted in Fig.~\ref{fig:prop}, $\bm{Q}_t$ undergoes self-attention with $\bm{Q}_R$ and completes cross-attention with image features $\bm{f}_t$. 
Therefore, it simultaneously integrates both the referring cues and the visual features from the current frame.
The updated $\bm{Q}_{t+1}$ then will serve as an input query of the next frame $\bm{I}_{t+1}$ after passing by a 3-layer MLP adapter.
By cross-frame referring propagation, TrackGPT obtains temporal priors that can adapt to changing scenarios, thus improving the accuracy of tracking.

\subsection{Optimization}
We fix the pre-trained vision encoder and finetune the mask decoder and MLP projection layers for TrackGPT training. 
For the LVLM brain, we employ LoRA~\cite{lora} to perform parameter-efficient fine-tuning.
We train TrackGPT end-to-end by a joint loss consisting of a text generation loss $\mathcal{L}_{text}$, a mask prediction loss $\mathcal{L}_{mask}$, and a purport score prediction loss $\mathcal{L}_{po}$.
The learning objective is formulated as:
\begin{align}
	\label{eq:locate_loss}
	\mathcal{L}_{joint} &= \lambda_{text}\mathcal{L}_{text} + \lambda_{mask}\mathcal{L}_{mask} + \lambda_{po}\mathcal{L}_{po}.
\end{align} 
where $\lambda_{text}$, $\lambda_{mask}$ and $\lambda_{po}$ are the corresponding loss coefficients. 
$\mathcal{L}_{text}$ is an auto-regressive cross-entropy loss,  $\lambda_{mask}$ is the mask-related loss that consists of binary mask cross-entropy loss and DICE~\cite{dice} loss. $\mathcal{L}_{po}$ is a MSE loss.
In our experiments, we set $\lambda_{text}=1$, $\lambda_{mask}=1$, and $\lambda_{po}=1$. For $\mathcal{L}_{mask}$, we set the weights for the cross-entropy loss and DICE loss to 2.0 and 0.5, respectively.

\subsection{InsTrack Benchmark}
To better evaluate and analyze the instruction tracking capability, 
we assemble an instruction tracking benchmark named InsTrack.
We collected a total of 143/40 video sequences for instruction tracking tuning and evaluation.
Each video is accompanied by 6 tracking instructions, resulting in over 1,000 instruction-video pairs in total.
Among them, the first instruction is manually provided by human annotators, and we take these instructions as seed instructions, the remaining ones are rephrased from the seed instruction with the assistance of Chat-GPT3.5~\cite{chatgpt}.
The video sequences are collected from two types of video object-tracking datasets.
One is the referring video object segmentation dataset (we select 32 sequences with a target object of interest from 90 sequences from Refer-DAVIS$_{17}$~\cite{refer-davis}), where we re-annotate the implicit instruction language for the object in the videos.
The other is a video instance segmentation dataset (we select 151 sequences from 3,536 sequences from VIPSeg~\cite{vigseg}). We provided tracking instructions for the target object of interest from the selected sequences. More details of InsTrack are provided in the {supplementary material}.

\section{Experiment}
\subsection{Implementation Details}

\begin{table*}[t]
	\centering
	\small
	\caption{{The quantitative evaluation on Refer-Youtube-VOS and Refer-DAVIS$_{17}$}, with region similarity $\mathcal{J}$, boundary accuracy $\mathcal{F}$, and average of $\mathcal{J}$\&$\mathcal{F}$. For online approaches, the best results are in \textbf{bold} font and the second ones are underlined.}
	\vspace{-3mm}
	\resizebox{0.90\textwidth}{!}{
		\setlength\tabcolsep{8pt}
		\renewcommand\arraystretch{1.0}
		\begin{tabular}{l||c|c|ccc|ccc}
			\hline
			\multirow{2}{*}{Method} & \multirow{2}{*}{Year} & \multirow{2}{*}{Online/Offline} & \multicolumn{3}{c|}{Refer-Youtube-VOS} &\multicolumn{3}{c}{Refer-DAVIS$_{17}$}  \\
			\cline{4-9}
			
			&& &$\mathcal{J}$\&$\mathcal{F}$& $\mathcal{J}$ & $\mathcal{F}$& $\mathcal{J}$\&$\mathcal{F}$ & $\mathcal{J}$ & $\mathcal{F}$ \\ 
			\hline
			CMPC-V~\cite{cmpc} &2021&offline &47.5 &45.6 &49.3 &-&-&- \\
			PMINet + CFBI~\cite{pminet}  &2021&offline  & 54.2 & 53.0 & 55.5 & - & - & - \\
			CITD~\cite{citd}  &2021& offline & 61.4 & 60.0 & 62.7  & - & - & -\\
			MLSA~\cite{mlsa} &2022& offline &49.7 & 48.4 & 50.9 & 57.9 & 53.8 & 62.0 \\
			MTTR~\cite{mttr}  &2022& offline & 55.3 & 54.0 & 56.6 & - & - & - \\
			ReferFormer~\cite{referformer}  &2022& offline & 62.9 & {61.3} & 64.6 & 61.1 &58.1 &64.1\\
			R$^2$-VOS~\cite{r2vos} &2023&offline &61.3 &59.6 &63.1 &-&-&- \\ 
			SgMg~\cite{sgmg} &2023&offline &65.7&63.9 &67.4&63.3&60.6 &66.0 \\
			OnlineRefer~\cite{onlinerefer} &2023& semi-online &{62.9}& 61.0 &{64.7}& {62.4}&{59.1}&{65.6} \\
			\hline
			CMSA~\cite{cmsa} &2019& online & 34.9  & 33.3 & 36.5 & 34.7  & 32.1 & 37.2\\
			URVOS~\cite{refer-youtube-vos}&2020 & online & 47.2 & 45.2 & 49.1  & 51.6&  47.2 & 55.9 \\
			YOFO~\cite{yofo} &2022 & online &48.6 &47.5 &49.7 &53.3 &48.8 &57.8 \\
			LBDT~\cite{lbdt} &2022& online &49.4 &48.2 &50.6 & 54.3 & -&- \\
			R$^2$-VOS~\cite{r2vos} &2023&online &\underline{60.2} &\underline{58.9} &\underline{61.5} &-&-&- \\ 
			OnlineRefer~\cite{onlinerefer} &2023& online &\textbf{63.5}&\textbf{61.6}&\textbf{65.5}&\underline{64.8}&\underline{61.6}&\underline{67.7} 	\\
			LISA-7B$^*$~\cite{lisa} &2023& online &50.2&49.7&50.6&58.4&54.9&61.9\\
			LISA-13B$^*$~\cite{lisa} &2023& online &52.6&52.1&53.0&60.7&56.8&64.6\\
			
			\hline
			TrackGPT-7B (ours)  &2023&online   &56.4  &55.3 &57.4 &63.2  &59.4  &67.0 \\
			TrackGPT-13B (ours) &2023 &online   &59.5  &58.1 &60.8 &\bf{66.5} &\bf{62.7} &\bf{70.4}\\
			\hline 
	\end{tabular} }
	\vspace{-5pt}
	\label{table:sota_refer}
\end{table*}

\noindent\textbf{Training.}
The training process of TrackGPT consists of three stages.
{1}) Pretrained on image datasets. Following LISA~\cite{lisa}, we selected semantic segmentation datasets~\cite{ade2k, coco-stuff, paco-lvis, partimagenet, pascal-part}, referring segmentation datasets~\cite{refercoco, refercocog}, the visual question answering dataset (LLaVA-Instruct-150k~\cite{llava}), and the reasoning image segmentation dataset (ReasonSeg~\cite{lisa}) for the image-level pretraining of our TrackGPT.
Rich image-level data can effectively alleviate the relative scarcity of video referring/instruction tracking data.
{2}) Training on referring video object segmentation dataset.
To enable TrackGPT to achieve the ability to segment objects in videos, we conduct the second stage of training on Refer-Youtube-VOS~\cite{refer-youtube-vos}.
We randomly sample 3 frames within a video to simulate the initial frame $\bm{I}_0$, the current frame $\bm{I}_t$, as well as the next frame $\bm{I}_{t+1}$.
Similar to constructing prompts for images in~\cite{lisa}, we devise prompts for video sequences to train TrackGPT. The template is as follows:
\textit{“\textbf{USER}: \texttt{<IMAGE>} Can you track the \{text description\} in the video? \textbf{ASSISTANT}: Sure, I track the object. $\langle \text{TK} \rangle$, $\langle \text{PO} \rangle$”}.
Where the \textit{\{text description\}} is from the text description provided by Refer-Youtube-VOS~\cite{refer-youtube-vos}.
Following existing works~\cite{referformer,onlinerefer},
we report the results of the second-stage model on Refer-Youtube-VOS~\cite{refer-youtube-vos} and Refer-DAVIS$_{17}$.
{3}) Instruction tuning on InsTrack training data.
To enhance TrackGPT's ability to understand human intent for reasoning tracking, we further performed instruction tuning on the constructed InsTrack dataset.
In experiments, we find that this training phase led to significant improvements.
We train our models with 4 80GB Tesla A100 GPUs with AdamW~\cite{adamw} optimizer and deepspeed~\cite{deepspeed} for training acceleration.
The total batch size is set to 16.
Each stage is trained for 10 epochs, with 500 steps per epoch for the first and second stages, and 100 steps for the instruction tuning stage.
All three training stages take about 3 days.

\noindent\textbf{Model.}
We adopt LLaVA-7B~\cite{llava} as the large vision-language model for TrackGPT-7B, and LLaVA-13B~\cite{llava} for TrackGPT-13B.
The vision encoder is a fixed pre-trained ViT-H~\cite{vit} from SAM~\cite{sam}.
Different from some offline referring video object segmentation approaches that take video clips as inputs, TrackGPT is an online model, when dealing with long and ongoing videos, TrackGPT has a constant demand for computational resources, and performing online tracking aligns better with real-world tracking scenarios.
During inference, the rethinking threshold $\tau$ is set to 0.5 times of initial purport score, which results in an average of 3.0\% frames for rethinking on InsTrack benchmark.

\begin{table*}[t]
	\centering
	\small
	\caption{{The quantitative evaluation on InsTrack. The best two results are in \textbf{bold} and underlined fonts. IT: instruction tuning.}}
	\vspace{-3mm}
	\resizebox{0.90\textwidth}{!}{
		\setlength\tabcolsep{8pt}
		\renewcommand\arraystretch{1.0}
		\begin{tabular}{l||c|c|c|ccccc}
			\hline
			\multirow{2}{*}{Method} & \multirow{2}{*}{Year} & \multirow{2}{*}{Online/Offline} &\multirow{2}{*}{\tabincell{c}{{Reasoning}\\{Ability}}} & \multicolumn{5}{c}{InsTrack} \\
			\cline{5-9}
			
			&&&&$\mathcal{J}$\&$\mathcal{F}$& $\mathcal{J}$ & $\mathcal{F}$ &$\mathcal{J\!\!-\!\!R}$    &$\mathcal{F\!\!-\!\!R}$     \\ 
			\hline
			ReferFormer~\cite{referformer}  &2022& offline &\no &24.0&22.9&25.2&24.7&23.1\\
			SgMg~\cite{sgmg} &2023&offline &\no &33.4&31.9&35.0&34.6&35.1 \\
			\hline
			R$^2$-VOS~\cite{r2vos} &2023&online &\no &24.7&21.1&28.2&21.6&27.4  \\ 
			OnlineRefer~\cite{onlinerefer} &2023& online&\no &36.8&33.8&39.9&38.6&40.0 	\\
			LISA-7B$^*$~\cite{lisa} &2023& online&\yes &39.7&38.5&40.9&37.0&37.5 	\\
			LISA-13B$^*$~\cite{lisa} &2023& online&\yes &45.2&44.8&45.7&47.4& 44.7	\\
			\hline
			TrackGPT-7B  &2023&online&\yes  &43.3&42.1&44.5&47.4&46.2 \\
			TrackGPT-13B &2023 &online&\yes  &\underline{50.5}&48.8&\underline{52.1}&54.1&\underline{58.0}\\
			\hline
			TrackGPT-7B (IT)  &2023&online&\yes  &49.2&\underline{48.9}&49.4&\underline{55.9}&54.9 \\
			TrackGPT-13B (IT) &2023 &online&\yes  &\bf{54.9}&\bf{52.8}&\bf{56.9}&\bf{59.4}&\bf{62.9}\\
			\hline 
	\end{tabular} }
	\vspace{-8pt}
	\label{table:sota_inst}
\end{table*}

\begin{table}[t]
	\centering
	\renewcommand\arraystretch{1.0}
	\resizebox{0.46\textwidth}{!}{
		\setlength{\tabcolsep}{2.75mm}{
			\begin{tabular}{l||ccccc}
				\hline
				Components & $\mathcal{J} \& \mathcal{F}$ & $\mathcal{J}$ & $\mathcal{F}$ &$\mathcal{J\!\!-\!\!R}$ &$\mathcal{F\!\!-\!\!R}$\\
				\hline
				& \multicolumn{5}{c}{Refer-DAVIS$_{17}$} \\
				\hline
				Baseline &60.0  &55.9  &64.1  &61.4  &67.1  \\
				+RP &61.5 (+1.5)  &57.7	&65.2	&64.0&69.6 \\
				+RP+RT &64.3 (+2.8) &60.3&68.3	&67.1	&73.6\\
				\hline
				& \multicolumn{5}{c}{InsTrack} \\
				\hline
				Baseline &41.5  &40.0  &42.1  &44.8  &43.3 \\
				+RP &42.7 (+1.2)  &41.5  &44.0  &46.8  &45.6\\
				+RP+RT &44.0 (+1.3) &42.8&45.2&48.0&47.0\\
				\hline
	\end{tabular}}}
	\vspace{-2mm}
	\centering
	\caption{{Impact of different components in TrackGPT.} RP: cross-frame referring propagation, RT: rethinking mechanism.}
	\label{tab:module_effectiveness}
	\vspace{-0.2cm}
\end{table}

\subsection{Results on Referring Tracking}
To verify the effectiveness of TrackGPT on conventional referring tracking tasks, we make a comparison with existing state-of-the-art approaches on two benchmarks of Refer-Youtube-VOS~\cite{refer-youtube-vos} and Refer-DAVIS$_{17}$~\cite{refer-davis} in Tab.~\ref{table:sota_refer}.

\noindent\textbf{Refer-Youtube-VOS.}
Refer-Youtube-VOS~\cite{refer-youtube-vos} is a large-scale benchmark that covers 3,978 videos with $\sim$15K language descriptions, and 202 video sequences are used for evaluation. 
As reported in Tab.~\ref{table:sota_refer}, 
TrackGPT obtains competitive performance compared with existing state-of-the-arts (\eg, R$^2$-VOS~\cite{r2vos} and OnlineRefer~\cite{onlinerefer}).
TrackGPT-13B achieves a $\mathcal{J}\&\mathcal{F}$ of 59.5, which ranks third in online approaches.
We also extend the image segmentation method LISA~\cite{lisa} with reasoning capability to the task of referring video segmentation, using the \texttt{<SEG>} token generated from the initial frame to segment subsequent frames.
TrackGPT-7B achieves a 55.3 $\mathcal{J}$ and 57.4 $\mathcal{F}$, outperforms LISA-7B$^*$~\cite{lisa} with improvements of 5.6\% and 6.8\%, even significantly outperforming LISA-13B$^*$~\cite{lisa}.

\noindent\textbf{Refer-DAVIS$_{17}$.}
Ref-DAVIS$_{17}$~\cite{refer-davis} benchmark is built upon DAVIS$_{17}$~\cite{davis} by providing the language description for a specific object in each video, the test set contains 30 videos.
As shown in Tab.~\ref{table:sota_refer}, TrackGPT-13B obtain 66.5 $\mathcal{J}\&\mathcal{F}$, which is the new state-of-the-art performance.
TrackGPT-13B outperforms the best online algorithm, OnlineRefer~\cite{onlinerefer}, by 1.7\% in terms of $\mathcal{J}\&\mathcal{F}$, and achieves a 3.2\% improvement over the best offline algorithm, SgMg~\cite{sgmg}.
Moreover, TrackGPT-7B and TrackGPT-13B outperform LISA-7B$^*$
and LISA-13B$^*$ by a significant margin.
The performance of TrackGPT on Ref-DAVIS$_{17}$
demonstrates its capability to handle conventional explicit description-based video segmentation tasks as well.

\subsection{Results on Instruction Tracking}
\noindent\textbf{InsTrack.}
An instruction tracking benchmark --- InsTrack is constructed for instruction tracking evaluation.
We test TrackGPT and some recent state-of-the-art methods.
The results are reported in Tab.~\ref{table:sota_inst}.
We adopt region similarity $\mathcal{J}$, contour accuracy $\mathcal{F}$, their average value $\mathcal{J}\&\mathcal{F}$, their recall scores $\mathcal{J\!\!-\!\!R}$ and $\mathcal{F\!\!-\!\!R}$ for performance evaluation.
Notably, TrackGPT achieves state-of-the-art performance across all metrics, particularly TrackGPT-13B, achieves the best performance of 54.9 in terms of $\mathcal{J}\&\mathcal{F}$.
TrackGPT also significantly exceeds previous state-of-the-arts for online (OnlineRefer~\cite{onlinerefer} with 36.8 $\mathcal{J}\&\mathcal{F}$) and offline (SgMg~\cite{sgmg} with 33.4 $\mathcal{J}\&\mathcal{F}$)  referring video object segmentation.
These algorithms encounter significant performance degradation on the InsTrack benchmark due to the lack of reasoning capability.
When equipped with reasoning capability, LISA$^*$
exhibits more advanced performance compared to other RVOS methods. LISA-13B$^*$ achieves a  $\mathcal{J}$ score of 44.8 and a $\mathcal{F}$ score of 45.7.
With an LVLM reasoning brain,
TrackGPT-13B achieved a good performance of 50.5 $\mathcal{J}\&\mathcal{F}$ without conducting instruction tuning on InsTrack training set. 
After instruction tuning on InsTrack, TrackGPT-7B and TrackGPT-13B further significantly increase by 5.9\% and 4.4\% in terms of $\mathcal{J}\&\mathcal{F}$.

\subsection{Exploration Studies}

In this section, we conduct ablution studies with TrackGPT-7B to demonstrate the effectiveness of each component.

\noindent\textbf{Effectiveness of Referring Propagation. }
As described in Sec.~\ref{sec:perception}, we propose cross-frame referring propagation to efficiently propagate temporal cues to the next frame.
As shown in Tab.~\ref{tab:module_effectiveness},
after applying referring propagation, the $\mathcal{J}$ and $\mathcal{F}$ scores on Refer-DAVIS$_{17}$ improve 1.8\% and 1.1\%, and the $\mathcal{J\!\!-\!\!R}$ and $\mathcal{F\!\!-\!\!R}$ scores improve 2.6\% and 2.5\%.
This implies that with referring propagation, 
TrackGPT can generate more accurate mask prediction.
We also arrive at consistent conclusions on the InsTrack benchmark where
$\mathcal{J}$ and $\mathcal{F}$ are boosted from 40.0/42.1 to 41.5/44.0.

\begin{figure*}[t]
	\centering
	\vspace{-4mm}
	\includegraphics[width=0.975\textwidth]{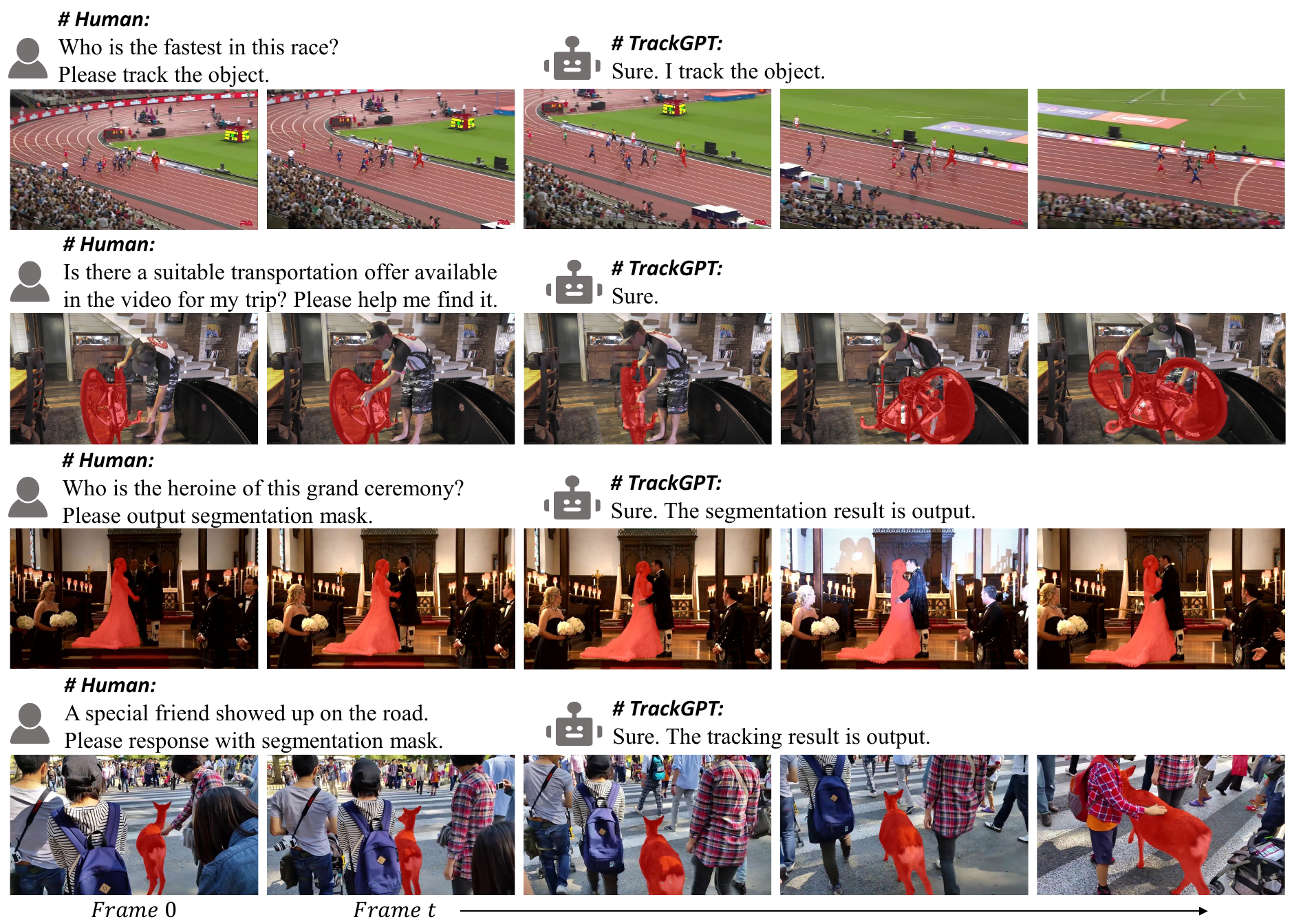}
	\vspace{-3mm}
	\caption{Quantitative results from InsTrack test set.
	TrackGPT comprehends the human instruction 
	and accurately tracks the target object.
    }
	\vspace{-3mm}
	\label{fig:vis}
\end{figure*}

\noindent\textbf{Effectiveness of Rethinking Mechanism.}
We propose a rethinking mechanism 
that when the current frame does not meet with the instruction purport, LVLM brain will 
update the $\langle \text{TK} \rangle$ token.
As shown in Tab.~\ref{tab:module_effectiveness}, after employing rethinking mechanism, on Refer-DAVIS$_{17}$~\cite{refer-davis}, $\mathcal{J}$ and $\mathcal{F}$ improve 2.6\% and 3.1\%, respectively.
On InsTrack, $\mathcal{J}\&\mathcal{F}$ is also improved from 42.7 to 44.0.
These results demonstrate the effectiveness of the rethinking mechanism.

\noindent\textbf{Importance of Pre-trained Vision Encoder.}
We adopt a pre-trained vision encoder from SAM~\cite{sam} for vision feature extraction.
As reported in Tab.~\ref{tab:train_ablution}, the comparison between experiments \ding{172} and \ding{173} demonstrates that 
it is imperative to utilize the encoder weights from SAM which is trained for segmentation at scale.
When replacing it with the ImageNet~\cite{imagenet} pre-trained weights, $\mathcal{J}\&\mathcal{F}$ decreases 5.6\%.

\noindent\textbf{Effectiveness of Pre-trained on Image Data.}
We first pre-train TrackGPT on image datasets to learn basic segment and conversation abilities.
As shown in experiment \ding{174} of Tab.~\ref{tab:train_ablution}, the performance is improved 4.4\% in terms of $\mathcal{J}\&\mathcal{F}$.
This result indicates that pre-training on image data can effectively alleviate the relative scarcity of data in referring video segmentation and instruction tracking.

\noindent\textbf{Effectiveness of Instruction Tuning. }
As shown in experiment \ding{175} in Tab.~\ref{tab:train_ablution}, after performing instruction tuning on the proposed InsTrack, the $\mathcal{J}\&\mathcal{F}$ score improves 4.1\%, and when we adopt Chat-GPT3.5~\cite{chatgpt} to rephrase instruction language to augment the instruction-video pairs (experiment \ding{176}), the $\mathcal{J}$ and $\mathcal{F}$ are boosted to 48.9 and 49.4.

\begin{table}[h]
	\centering
	\vspace{-2mm}
	\fontsize{8}{10}\selectfont  
	\setlength{\tabcolsep}{1.6mm}{
		\resizebox{\linewidth}{!}{
			\setlength{\tabcolsep}{0.5mm}{
				\begin{tabular}{c|cccc|ccc}
					\hline
					\multirow{2}{*}{\# ID} &
					\multirow{2}{*}{\tabincell{c}{{Pre-trained}\\{Vision Encoder}}} &
					\multirow{2}{*}{\tabincell{c}{{Pre-trained}\\{on Image Data}}} &
					\multirow{2}{*}{\tabincell{c}{{Instruction-tuned}\\{on InsTrack}}} &
					\multirow{2}{*}{\tabincell{c}{{Instruction}\\{Rephrasing}}} &
					\multicolumn{3}{c}{InsTrack}  \\
					\cline{6-8}
					&&&&& $\mathcal{J} \& \mathcal{F}$ & $\mathcal{J}$ & $\mathcal{F}$ \\
					\hline
					\ding{172} & &&&&33.3&30.0&36.7 \\
					\ding{173} &\yes &&&&38.9&37.2&40.4 \\
					\ding{174} &\yes&\yes& &&43.3&42.1&44.5 \\
					\ding{175} &\yes&\yes&\yes& &47.4 &45.3&49.5 \\
					\ding{176} &\yes&\yes&\yes&\yes&49.2&48.9&49.4 \\
					\hline
	\end{tabular}}}}
	\vspace{-3.mm}
	\caption{Ablation studies of training manners on InsTrack.
	}
	\vspace{-5.5mm}
	\label{tab:train_ablution}
\end{table}

\subsection{Quantitative Results}
As shown in Fig.~\ref{fig:vis}, we provide some typical quantitative results from TrackGPT on InsTrack.
Given a implied tracking instruction, \eg, “\textit{Who is the fastest in this race?}”
TrackGPT automatically reasons that it needs to find the athletes on the field and analyze who is running at the front and track him in subsequent video frames.
In other scenes, the instructions also cast queries that require world knowledge and complex reasoning, but TrackGPT performs well even if the target objects face various challenges \eg, appearance variation, occlusion, and visually-similar distractors.

\section{Conclusion}
This work proposes a new task, \ie, instruction tracking, in which only an implied tracking instruction is provided, and the tracker is responsible for understanding the human intent and completing the object tracking.
To tackle this challenge, we propose TrackGPT, which introduces an LVLM to serve as the brain for reasoning the instruction.
Besides, a rethinking mechanism and a cross-frame propagation module are designed to further improve the accuracy and robustness of TrackGPT. 
Moreover, we construct an instruction tracking benchmark named InsTrack for instruction tuning and evaluation.
Experiments demonstrate TrackGPT achieves promising performance on both conventional referring tracking and instruction tracking tasks.
We hope this work could catalyze more compelling research on utilizing the capabilities of LLMs for object tracking.

\noindent\textbf{Limitations.} 
Despite achieving promising performance on instruction tracking, TrackGPT is not able to handle the cases in a single round of inference when the human needs to track multiple objects.
We will investigate instruction tracking multiple objects in a video in future work.

{
	\small
	\bibliographystyle{ieeenat_fullname}
	\bibliography{main}
}

\end{document}